\documentclass[runningheads]{llncs}
\usepackage{graphicx}
\usepackage{amsmath,amssymb} 
\usepackage{color}
\usepackage{multirow}
\usepackage[width=122mm,left=12mm,paperwidth=146mm,height=193mm,top=12mm,paperheight=217mm]{geometry}
\begin{document}
\pagestyle{headings}
\mainmatter
\def\ECCV16SubNumber{***}  

\title{Examining Representational Similarity in ConvNets and the Primate Visual Cortex} 
\titlerunning{Examining Rep. Sim. in ConvNets and the Vis. Cortex}
\authorrunning{A. Dubey, Jayadeva and S. Agarwal}
\author{Abhimanyu Dubey, Jayadeva and Sumeet Agarwal}
\institute{\{ee2110061, jayadeva, sumeet\}@iitd.ac.in \\Indian Institute of Technology Delhi}
\maketitle
\begin{abstract}
We compare several ConvNets with different depth and regularization techniques with multi-unit macaque IT cortex recordings and assess the impact of the same on representational similarity with the primate visual cortex. We find that with increasing depth and validation performance, ConvNet features are closer to cortical IT representations.
\end{abstract}
\section{Introduction}
With the success of convolutional neural networks in object image classification \cite{krizhevsky2012imagenet,simonyan2014very,szegedy2015going,he2015deep}, it is of interest to examine how similar the representations learnt by these networks are to the visual representations learnt by the human brain \cite{yamins2014performance,gucclu2015deep,cichy2016}. In Cadieu {\em et al.}'s pioneering work on understanding representational similarity \cite{cadieu2014deep}, a comparison of the pre-final layer activations of deep neural net models with multi-unit IT cortex responses is presented, which confirms a significant correspondence between the two representations for the task of core object recognition. We aim to shed more light on the role of engineered aspects of these deep networks, examining the effect of variation in regularization, network depth and model size on the representational similarity to multi-unit IT responses and classification accuracy. This can help understand whether sparsity and increased network depth create representations which are closer to representations employed by the primate visual cortex, and also validate the hypothesis that the mammalian visual cortex is the best known object detector, by checking whether increased proximity to cortical representations implies an increase in recognition performance.
\section{Methodology}
For each ConvNet, we supply input images from $n$ output classes and perform a feed-forward operation to obtain the layer-wise activations, and extract the activation at the penultimate fully-connected layer. For any network architecture $N$, we denote the set of activations for the image set by $\mathbf{A}_N$. For any particular target class $i \in [1,n]$, the set of activations is given by $\mathbf{A}^i_N$. For any set of activations, we can define the \textit{average activation} as $\phi(\mathbf{A}^i_N) = (\sum_{a \in \mathbf{A}^i_N} a)/(|\mathbf{A}^i_N|)$. \\ 
\hspace*{10pt} Given these representations, we can compute the \textit{representational dissimilarity matrix} $\mathbf{R}_N$ as (\cite{cadieu2014deep})
\begin{align*}
\mathbf{R}_N = (r_{ij}) = 1 - \frac{\sigma(\phi(\mathbf{A}^i_N),\phi(\mathbf{A}^j_N))}{\sqrt{\sigma^2(\phi(\mathbf{A}^i_N)) \times \sigma^2(\phi(\mathbf{A}^j_N))}} ; \mathbf{R}_N \in \mathbb{R}^{n \times n}
\end{align*}
\hspace*{10pt} Where $\sigma$ and $\sigma^2$ denote covariance and variance respectively. We compute this matrix for the cortical responses as well, treating the responses as the set of activations as $\mathbf{R}_{IT}$. To measure representational similarity, we use ``similarity to IT dissimilarity matrix" ($s_{IT}$) as the default metric \cite{cadieu2014deep}, which is as the Spearman's rank correlation between the upper triangular, non-diagonal elements of the two matrices $\mathbf{R}_N$ and $\mathbf{R}_{IT}$. In essence, this metric would be expected to encapsulate the nature of variation across classes for each representation \cite{cadieu2014deep}.\\
\hspace*{10pt} To create the image set for activations, we randomly sample images from the total set in a manner identical to \cite{cadieu2014deep}. We also add noise to the ConvNet activations in order to account for the measurement noise present in the IT cortical responses following the experimental noise matched model in \cite{cadieu2014deep}.
\paragraph*{\textbf{Dataset}}
\begin{figure}[t]
\centering
\begin{minipage}[c]{0.75\textwidth}
\includegraphics[width=\textwidth]{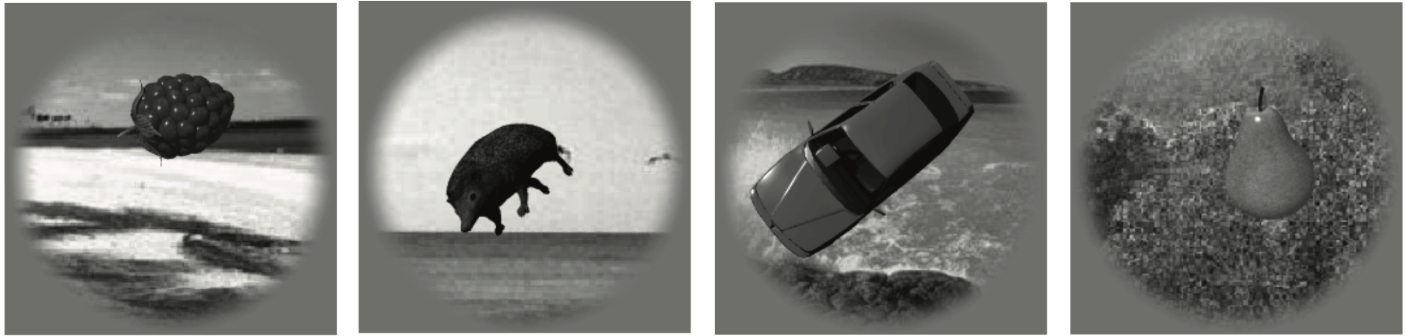}
\end{minipage}
\begin{minipage}[c]{0.22\textwidth}
\tiny
\caption{(Reproduced from \cite{cadieu2014deep}.) Samples from the dataset introduced by \cite{cadieu2014deep}.\label{dataset}}
\end{minipage}
\end{figure}
Cadieu {\em et al.} \cite{cadieu2014deep} introduce a dataset of multi-unit V4 and IT cortex responses across two male rhesus monkeys while they were presented synthetically generated object image samples. These recordings were taken from the 128 most visually driven neural measurement sites determined via a separate pilot dataset. We employ these recordings as the inferior-temporal (IT) representations (features) for our comparative experiments, and we have a total of 1,960 images for 7 categories (see Figure \ref{dataset}).
\paragraph*{\textbf{Evaluation}}
We evaluate the best performing ConvNet architectures on ImageNet LSVRC and employ their publicly available pre-trained weights. We experiment on both the original architectures and retrained architectures using different regularization schemes - L1, L2 (default) and DeCov \cite{cogswell2015reducing}. DeCov tries to learn independent, generalizable filters by minimizing the covariances of filter activations, with the additional \textsc{DeCov} loss $\mathbf{\mathcal{L}}_{DeCov}$ at each hidden layer given by (\cite{cogswell2015reducing})
\begin{align*}
\mathbf{\mathcal{L}}_{DeCov} &= \frac{1}{2} (\lVert C \rVert^2_F - \lVert diag(C) \rVert^2_2) \\ C = (c_{i,j}) &= \frac{1}{|\mathbf{A}^{batch}_N|}\sum_k (h^k_i - \mu_i)(h^k_j - \mu_j)
\end{align*} The $diag(\cdot)$ operator extracts the main diagonal of a matrix into a vector. The function $C$ is the matrix of covariances of all pairs of activations $h^k_i,h^k_j$ at a hidden layer $k$, and $\mathbf{A}^{batch}_N \subset \mathbf{A}_N$ is the sample batch during training of architecture $N$, and $\mu_i = \frac{1}{|\mathbf{A}^{batch}_N|} \sum_k h^k_i$ is the sample mean of activation $i$ over the batch. We add this loss to each layer of the \textsc{DeCov} network \cite{cogswell2015reducing} and tune the hyperparameter via cross-validation. For L1, we replace weight-decay (L2) with L1-norm, and tune via cross-validation. We maintain the original weight decay formulation and weights for L2. All networks are trained with dropout.
\section{Results and Conclusions}
\begin{figure}[t]
\centering
\includegraphics[width=\textwidth]{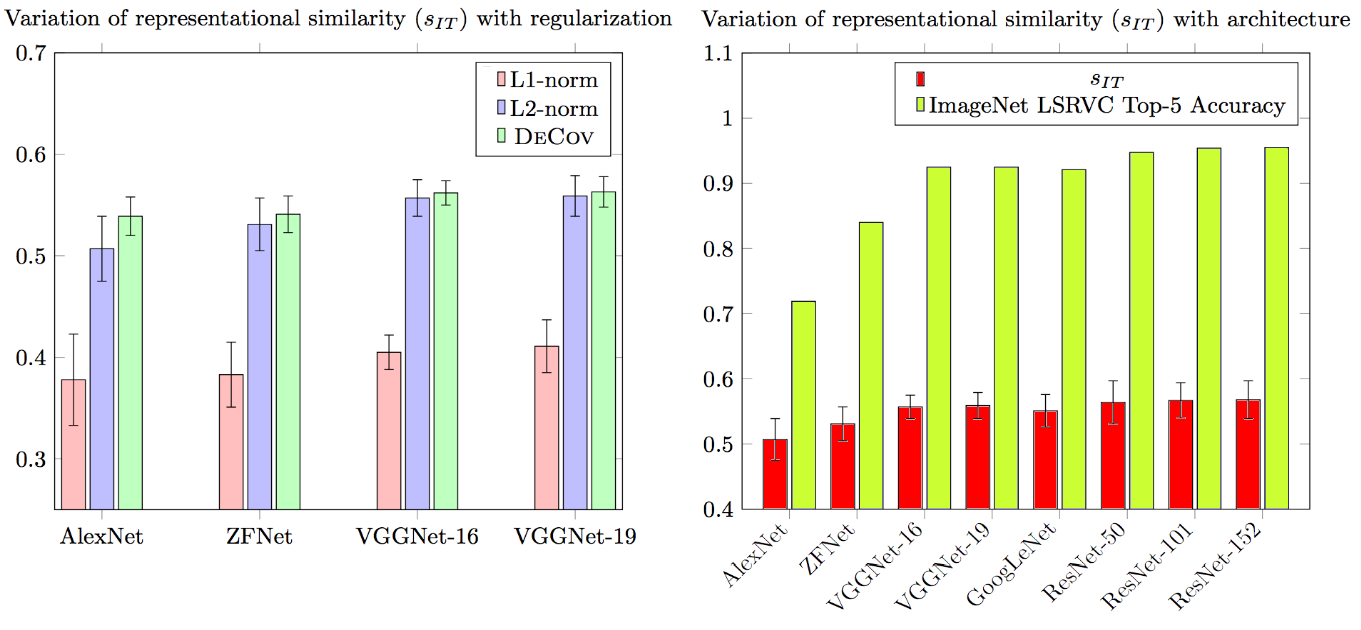}
\caption{Variation of representational similarity $s_{IT}$ with model regularization (left) and validation performance (right). We see that representational similarity consistently increases with validation performance across architectures.\label{results}}
\end{figure}
\hspace*{10pt} While Cadieu {\em et al.} \cite{cadieu2014deep} validated the plausibility of deep neural net representations being similar to cortical representations, our work seeks to examine a wider range of variation in the characteristics of the deep learning models used. Our results show that by increasing network depth, we observe a marginal increase in representational similarity, which is also consistent with an increase in validation accuracy. Another interesting result is that by decorrelating representations, we observe a higher representational similarity, even though validation performance is similar.
\begin{table}[t!]
\centering
\begin{minipage}[c]{0.71\textwidth}
\tiny
\begin{tabular}{|c|c|c|c|c|}
\hline 
\textbf{Deep} & \multirow{2}{*}{\textbf{Layers}} & \multirow{2}{*}{\textbf{$s_{IT}$}} & \textbf{Accuracy} & \textbf{ImageNet Top-5} \\ 
\textbf{Network}&&&\textbf{on \cite{cadieu2014deep}}&\textbf{Error Rate}\\ \hline
\multicolumn{5}{|c|}{\textit{Original Architectures}} \\ \hline
AlexNet \cite{krizhevsky2012imagenet}& 8 & 0.507 ($\pm 0.032$) & 0.523 ($\pm 0.025$)& 18.1\% \\
ZFNet \cite{zeiler2014visualizing}& 8 & 0.531 ($\pm 0.026$) & 0.568 ($\pm 0.019$)& 16.0\% \\
VGGNet-16 \cite{simonyan2014very}& 16 & 0.557 ($\pm 0.018$) & 0.580 ($\pm 0.021$)& 7.5\% \\
VGGNet-19 \cite{simonyan2014very}& 19 & 0.559 ($\pm 0.020$) & 0.582 ($\pm 0.023$)& 7.5\% \\
GoogLeNet \cite{szegedy2015going}& 22 & 0.551 ($\pm 0.025$) & 0.575 ($\pm 0.021$)& 7.89\% \\
ResNet-50 \cite{he2015deep}& 50 & 0.564 ($\pm 0.033$) & 0.601 ($\pm 0.011$)& 5.25\% \\
ResNet-101 \cite{he2015deep}& 101 & 0.567 ($\pm 0.027$) & 0.603 ($\pm 0.010$)& 4.60\% \\
ResNet-152 \cite{he2015deep}& 152 & 0.568 ($\pm 0.029$) & 0.612 ($\pm 0.013$)& 4.49\% \\
\hline
\multicolumn{5}{|c|}{\textit{Modified Architectures}} \\ \hline
AlexNet-L1\cite{krizhevsky2012imagenet}& 8 & 0.378 ($\pm 0.045$) & 0.413 ($\pm 0.019$)& 27.2\% \\
ZFNet-L1\cite{krizhevsky2012imagenet}& 8 & 0.383 ($\pm 0.032$) & 0.431 ($\pm 0.017$)& 23.9\% \\
VGGNet-16-L1\cite{krizhevsky2012imagenet}& 16 & 0.405 ($\pm 0.017$) & 0.434 ($\pm 0.029$)& 13.5\% \\
VGGNet-19-L1\cite{krizhevsky2012imagenet}& 19 & 0.411 ($\pm 0.026$) & 0.437 ($\pm 0.015$)& 12.9\% \\
AlexNet-DeCov \cite{cogswell2015reducing} & 8 & 0.539 ($\pm 0.019$) & 0.521 ($\pm 0.027)$ & 20.0\% \\
ZFNet-DeCov \cite{cogswell2015reducing} & 8 & 0.541 ($\pm 0.018$) & 0.528 ($\pm 0.025)$ & 18.8\% \\
VGGNet-16-DeCov \cite{cogswell2015reducing} & 16 & 0.562 ($\pm 0.012$) & 0.556 ($\pm 0.019)$ & 11.6\% \\
VGGNet-19-DeCov \cite{cogswell2015reducing} & 19 & 0.563 ($\pm 0.015$) & 0.561 ($\pm 0.024)$ & 11.2\% \\
\hline
\end{tabular}
\end{minipage}
\begin{minipage}[c]{0.28\textwidth}
\caption{A comparison of the representational similarity of ConvNets to the IT cortex ($s_{IT}$), as per \cite{cadieu2014deep}. Original accuracies on ImageNet are reported directly. Accuracy on the data set of \cite{cadieu2014deep} is obtained using a linear SVM trained on pre-final activations.}
\end{minipage}
\label{tab:main}
\end{table}
Our experiments offer a preliminary understanding of the effect of network depth and model complexity control on the similarity between deep neural net and cortical representations. Such an approach can help provide some pointers to the learning architectures and mechanisms employed by the brain. Substantial further work needs to be done: comparison of lower-layer activations of the network with preliminary regions of object recognition (V1, V2), visualization of the effect of context on representational similarity and understanding the impact of dropout and structural risk minimization on networks are some extensions we have initiated. The ultimate aim is to create models similar to humans in both representation and performance on complex vision tasks, as a means of better understanding, or `reverse engineering', the human visual system itself.
\bibliographystyle{splncs}
\small
\bibliography{egbib}
\end{document}